\documentclass[10pt,twocolumn,letterpaper]{article}

\usepackage{iccv}
\usepackage{times}
\usepackage{epsfig}
\usepackage{graphicx}
\usepackage{caption}
\usepackage{amsmath}
\usepackage{siunitx}
\usepackage{amssymb}
\usepackage{mwe}
\usepackage{times}
\usepackage{epsfig}
\usepackage{graphicx}
\usepackage{amsmath}
\usepackage{amssymb}
\usepackage{helvet}   
\usepackage{courier}  
\usepackage{url}      
\usepackage{xcolor}
\usepackage[accsupp]{axessibility} 
\usepackage{pifont}
\usepackage{booktabs}
\usepackage{multirow}
\usepackage{subcaption}
\usepackage{algorithmic}
\usepackage[linesnumbered,ruled,vlined]{algorithm2e}
\usepackage{booktabs}
\usepackage{float}
\usepackage{lipsum}
\usepackage[breaklinks=true,colorlinks,bookmarks=false]{hyperref}

\iccvfinalcopy 

\def\iccvPaperID{6415} 

\ificcvfinal\fi

\begin{document}

\title{TEMPO: Efficient Multi-View Pose Estimation, Tracking, and Forecasting}

\author{Rohan Choudhury \quad Kris M. Kitani
\quad L\'{a}szl\'{o} A. Jeni
\\
Robotics Institute, Carnegie Mellon University\\
{\tt\small {\{rchoudhu, kmkitani\}@andrew.cmu.edu}} \quad {\tt\small laszlojeni@cmu.edu} \\
}


\twocolumn[{%
\renewcommand\twocolumn[1][]{#1}%
\maketitle
\begin{center}
\centering
\captionsetup{type=figure}
\resizebox{0.97\textwidth}{!}{\includegraphics[width=\linewidth]{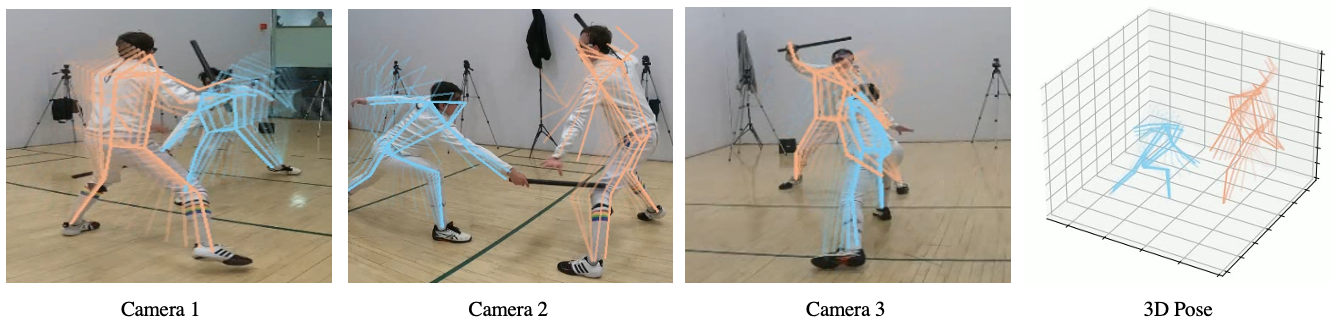}}
\captionof{figure}{We propose \textbf{TEMPO}: TEMporal POse Estimation, a method for multi-view, multi-person pose estimation, tracking and forecasting. TEMPO uses a recurrent architecture to learn a spatiotemporal representation, significantly improving the pose estimation accuracy while preserving speed at inference time. In each view, we show TEMPO's predicted pose skeletons over 10 frames, colored by tracker identity. Lighter colors correspond to previous frames.}
\label{fig:teaser}
\end{center}
}]
\ificcvfinal\thispagestyle{empty}\fi

\begin{abstract}
\vspace{-1.45em}
Existing volumetric methods for predicting 3D human pose estimation are accurate, but computationally expensive and optimized for single time-step prediction. We present TEMPO, an efficient multi-view pose estimation model that learns a robust spatiotemporal representation, improving pose accuracy while also tracking and forecasting human pose. We significantly reduce computation compared to the state-of-the-art by recurrently computing per-person 2D pose features, fusing both spatial and temporal information into a single representation. In doing so, our model is able to use spatiotemporal context to predict more accurate human poses without sacrificing efficiency. We further use this representation to track human poses over time as well as predict future poses. Finally, we demonstrate that our model is able to generalize across datasets without scene-specific fine-tuning. TEMPO achieves 10$\%$ better MPJPE with a 33$\times$ improvement in FPS compared to TesseTrack on the challenging CMU Panoptic Studio dataset. Our code and demos are available at \url{https://rccchoudhury.github.io/tempo2023/}.
\end{abstract}

\vspace{-1em}
\section{Introduction}
Estimating the pose of several people from multiple overlapping cameras is a crucial vision problem. Volumetric multi-view methods, which lift 2D image features from each camera view to a feature volume then regress 3D pose, are currently the state of the art \cite{voxelpose, tessetrack, fastervoxelpose, iskakov2019learnable} in this task.
 These approaches produce significantly more accurate poses than geometric alternatives, but suffer from two key limitations. First, the most accurate methods use either 3D convolutions \cite{voxelpose, tessetrack, zhang2022voxeltrack} or cross-view transformers \cite{wang2021mvp} which are slow and prevent real-time inference. Secondly, most methods are designed for estimating pose at a single timestep and are unable to reason over time, limiting their accuracy and preventing their use for tasks like motion prediction.
 
\begin{figure*}
\begin{center}
\includegraphics[height=3in, width=\linewidth]{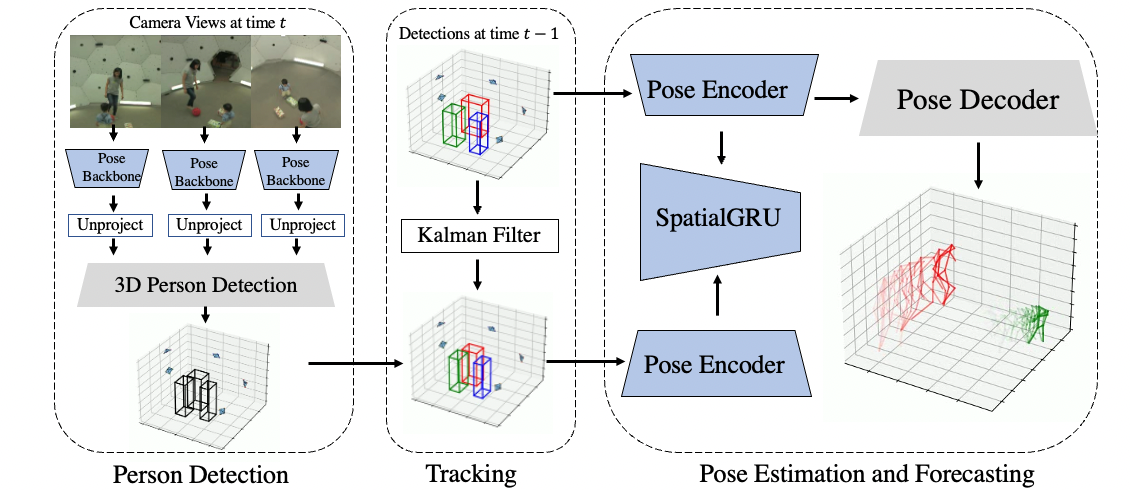}
\end{center}
\caption{The overall model architecture. We begin by (1) extracting features from each image with the backbone network and unprojecting those features to a 3D volume. In step (2), we use the volume to detect each person in the scene, and (3) associate the detections from the current timestep to the previous one. We then (4) fuse the features from each person with our temporal model and produce a final pose estimate.}
\label{fig:main_arch}
\end{figure*}

We propose TEMPO, a multi-view \underline{TEM}poral \underline{PO}se estimation method that addresses both of these issues. TEMPO uses \textit{temporal context} from previous timesteps to produce smoother and more accurate pose estimates. Our model tracks people over time, predicts future pose and runs efficiently, achieving near real-time performance on existing benchmarks.
The key insight behind TEMPO, inspired by work in 3D object detection \cite{bevformer, fiery}, is that recurrently aggregating spatiotemporal context results in powerful learned representations while being computationally efficient. To do this, we decompose the problem into three stages, illustrated in Figure \ref{fig:main_arch}. Given an input RGB video from multiple static, calibrated cameras, at a given timestep $t$ we first detect the locations of each person in the scene by unprojecting image features from each view to a common 3D volume. We then regress 3D bounding boxes centered on each person, and perform tracking by matching the box centers with the detections from the previous timestep $t-1$. For each detected person, we compute a spatiotemporal pose representation by recurrently combining features from current and previous timesteps. We then decode the representation into an estimate of the current pose as well as poses at future timesteps. Unlike existing work \cite{voxelpose, fastervoxelpose, tessetrack, zhang2022voxeltrack}, our method is able to perform temporal tasks like tracking and forecasting without sacrificing efficiency.
 
We evaluate our method on several pose estimation benchmarks. TEMPO achieves state of the art results on the challenging CMU Panoptic Studio dataset \cite{joo_iccv_2015} by 10$\%$, and is competitive on the Campus, Shelf and Human3.6M datasets. We additionally collect our own multi-view dataset consisting of highly dynamic scenes, on which TEMPO achieves the best result by a large margin. We show that our model achieves competitive results in pose tracking and evaluate the pose forecasting quality on the CMU Panoptic dataset. Additionally, multi-view pose estimation methods are almost always evaluated on the same dataset they are trained on, leading to results that are specific to certain scenes and camera configurations. We measure our method's ability to generalize across different datasets and find that our method can transfer without additional fine tuning.
To summarize, our key contributions are that:
\begin{itemize}
\setlength\itemsep{-0.2em}
\item We develop the most accurate multi-view, multi-person 3D human pose estimation model. Our model uses temporal context to produce smoother and more accurate poses.
\item Our model runs efficiently with no performance degradation.
\item Our model tracks and forecasts human pose for every person in the scene.
\item We evaluate the generalization of our model across multiple datasets and camera configurations.
\end{itemize}

\section{Related Work}
\textbf{3D Pose Estimation, Tracking and Forecasting}Approaches for recovering and tracking 3D human pose are usually limited to monocular video. Such methods for pose estimation \cite{sun2021monocular, bogo2016keep, kanazawa2017end, kocabas2020vibe} and pose tracking \cite{ rajasegaran2022tracking, rajasegaran2021tracking} are highly efficient, but perform significantly worse than multi-view methods in 3D pose estimation accuracy due to the inherent ambiguity of monocular input.

Furthermore, methods in human pose forecasting \cite{mao2019learning} usually predict future motion from ground truth pose histories. Our approach follows \cite{cao2020long, snipper} and predicts pose directly from a sequence of video frames. Snipper \cite{snipper} is the closest to our method and uses a spatiotemporal transformer to jointly estimate, track and forecast pose from a monocular video. Our method differs in that it is able to produce highly accurate estimates using multi-view information while running efficiently.

\textbf{Multi-View Pose Estimation}
Early work in multi-view human pose estimation was limited to the single-person case, with \cite{belagiannis_campus, kostrikov2014depth, pavlakos2017harvesting, amin2013multi} using pictorial structure models to improve over basic triangulation. More recent approaches \cite{he2020epipolar, qiu2019cross, iskakov2019learnable, bultmann2021real, kadkhodamohammadi2021generalizable} improve this result by using advanced deep architectures like 3D CNNs and transformers, and others \cite{chen2022structural, ci2019optimizing} introduce priors on human shape for additional performance.
Our method is most similar to \cite{iskakov2019learnable}, which uses 3D CNNs to regress pose directly from a feature volume. We also follow \cite{iskakov2019learnable, kadkhodamohammadi2021generalizable} in analyzing our model's transfer to multiple datasets, extending their qualitative, single-person analysis to a quantitative measurement of performance on several multi-person datasets.

In the multi-person setting, early approaches like \cite{belagiannis_campus, dong2019fast, dong2021fastpami, Xu_2022_BMVC, bridgeman2019multi} associate 2D pose estimates from each view, then fuse the matched 2D poses into 3D. 
Other methods aimed towards multi-person motion capture use Re-ID features \cite{dong2019fast, Xu_2022_BMVC}, 4D graph cuts \cite{zhang20204d}, plane sweep stereo \cite{planesweeppose}, cross-view graph matching \cite{wu2021graph}, or optimize SMPL parameters \cite{dong2021shape} to produce 3D poses from 2D pose estimates in each view. These methods can generalize across the data sources, but typically have much less accurate predictions compared to \textit{volumetric} methods. These first unproject learned 2D image features into a 3D volume and regress pose directly from the 3D features with neural networks. Both  \cite{iskakov2019learnable} and \cite{voxelpose} use computationally expensive 3D CNNs for the pose estimation step.
Follow up work includes Faster VoxelPose \cite{fastervoxelpose}which  replaces these 3D CNNs with 2D CNNs for a large speedup, and TesseTrack \cite{tessetrack}, which uses 4D CNNs to reason over multiple timesteps. Our method combines the best of both: we efficiently incorporate spatiotemporal information with only 2D CNNs and a lightweight recurrent network.

\textbf{3D Object Detection and Forecasting} 
Historically, 3D object detection and instance segmentation methods for autonomous driving have led development in using multi-view images. One key similarity to our work is the aggregation of 2D image features into a single 3D volume. While \cite{sitzmann2019deepvoxels, tung2019learning} use the same bilinear unprojection strategy as our method, several works \cite{philion2020lift, harley2022simple, bevformer} propose alternatives such as predicting per-pixel depth.
Other works also use temporal information for detection for tracking objects through occlusion and spotting hard-to-see objects; \cite{fiery, bevformer, timewilltell} concretely demonstrate the benefits of incorporating spatiotemporal context. In particular, FIERY \cite{fiery} uses temporal information for future instance prediction and BEVFormer \cite{bevformer} efficiently aggregates temporal information with a recurrent architecture, both of which inspired our method. Furthermore, \cite{janai2018unsupervised, harley2022particle} use per-timestep supervision to track pixels through occlusion, an idea which we adapt for reasoning about human pose over multiple frames.

\section{Method}
Our method assumes access to calibrated time-synchronized videos from one or more cameras. At training time, we assume access to $T$ sets of $N$ RGB images from different cameras, while at inference time, we have a single set of $N$ images corresponding to the current timestep. In order to enable TEMPO to transfer to new camera configurations and settings, we compute the dimensions of the space and size of the voxel volume directly from the camera matrices. We set the height of the volume to a constant \SI{2}{\meter}, while setting the length and width of the volume to be the bounding box of the camera extrinsics from a top-down view, and center the space at the mean of the camera locations.  

\subsection{Preliminaries}
We briefly review the person detection and pose estimation modules used by VoxelPose \cite{voxelpose}, Tessetrack \cite{tessetrack}, and Faster VoxelPose \cite{fastervoxelpose} that TEMPO builds upon. We refer the reader to the original papers for further detail.

\subsubsection{Person Detection}
The detection module aims to estimate the location of the \textit{root joint} as well as tight 3D bounding box for each person in the scene. Following previous work, we define the root joint as the mid-hip.
 At a given time $t$, the detector module takes as input a set of $N$ images, each corresponding to a different camera view of the same scene at time $t$.
For each image, we  extract features with a pretrained backbone, resulting in $N$ feature maps $\textbf{F}^t_1, \textbf{F}^t_2, \hdots, \textbf{F}^t_N$. 

Given the camera matrices for each view $\mathbf{C}^t_1, \mathbf{C}^t_2, \hdots, \mathbf{C}^t_N$, we use the bilinear sampling procedure from \cite{iskakov2019learnable, voxelpose, harley2022simple}. For a voxel $v \in V$ with coordinates $\mathbf{x}$, we have
\begin{equation}
    v = \sum_{i=1}^N \mathbf{F}^t_i(\mathbf{C}_i\mathbf{x})
\end{equation}
where $\textbf{F}^t_i(\textbf{x})$ is the feature map $\textbf{F}^t_i$'s value at position $\textbf{x}$, obtained by bilinear sampling.
We then compute a birds-eye view representation of $V$ by taking the maximum along the $z$-axis:
\begin{equation}
    \textbf{F}^t_{\mathrm{BEV}} = \max_{z} V
\end{equation}
We use a 2D CNN to produce a 2D heatmap of $\mathbf{H}^t$ from $\textbf{F}^t_{\mathrm{BEV}}$
of the $(x, y)$ locations of every root joint in the scene.
We then  sample the top $K$ locations from $\mathbf{H}^t$, yielding proposals $(x_1, y_1), (x_2, y_2), \hdots (x_K, y_K)$. For each proposal location, we obtain the corresponding feature column $V|_{x, y}$ and apply a 1D CNN to regress a 1D heatmap of the root joint's height, denoted $\mathbf{H}^t_k$. We then sample the maximum $z$ coordinate for each $\mathbf{H}^t_k$, and combine these to produce a set of detections $D_t = \{(x_1, y_1, z_1), \hdots (x_K, y_k, z_K)\}$. Finally, we regress width, length and centerness from  $\textbf{F}^t_{\mathrm{BEV}}$ with a multi-headed 2D CNN to produce bounding box predictions for each proposal.
The loss function for the detection module has three terms. First, $L_{2D}$ is the distance between the predicted 2D heatmap and the ground truth, given by 
\begin{figure*}
    \centering
    \includegraphics[width=\linewidth]{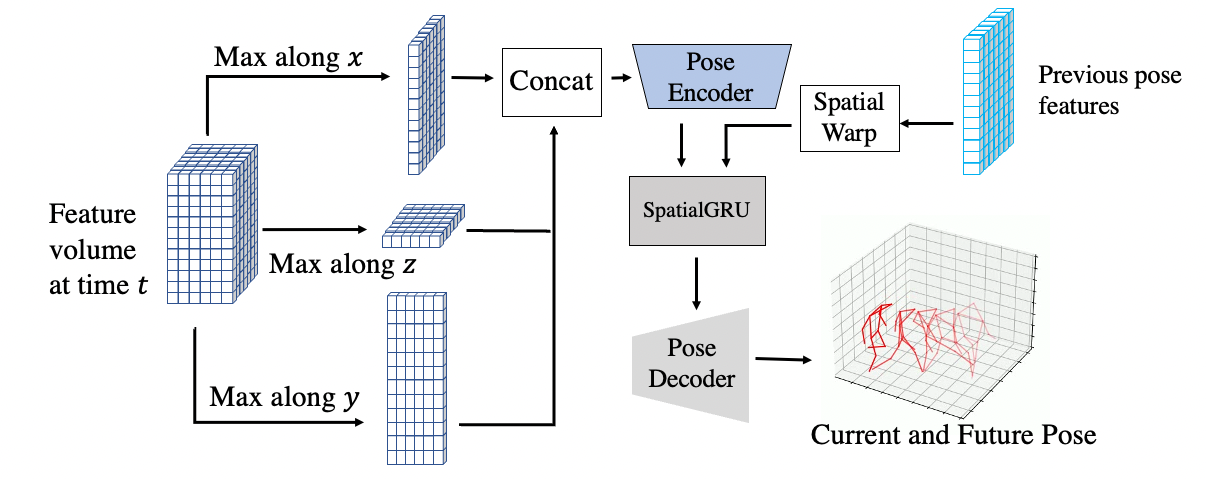}
    \caption{A closer look at the temporal representation used by our model. Following \cite{fastervoxelpose}, we first project the feature volume to each of the three planes, and concatenate the projections channel-wise. We pass this feature map through an encoder network. We use this feature encoding as input to the SpatialGRU module, using the spatially warped pose feature from the previous timestep as a hidden state. We use the SpatialGRU module to produce features at the current and future timesteps, which we decode into human poses with the pose decoder network.}
    \label{fig:temporal_rep_figure}
\end{figure*}

\begin{equation}
    L_{2D} = \sum_{t=1}^T \sum_{(x, y))} \lVert \textbf{H}^t(x, y) - \textbf{H}^t_{GT}(x, y)\rVert
\end{equation}
We also compute the loss on the predicted 1D heatmap:
\begin{equation}
    L_{1D} = \sum_{t=1}^T \sum_{k=1}^K \sum_{z} \lVert \textbf{H}^t_k(z) - \textbf{H}^t_{k, GT}(z) \rVert
\end{equation}
Finally, we include the bounding box regression loss
\begin{align}
L_{\mathrm{bbox}} = \sum_{t=1}^T\sum_{(i, j) \in U} \lVert \mathbf{S}(i, j) - \mathbf{S}_{GT}(i, j) \rVert_1 
\end{align}
The total detection loss is the sum of the above terms, with 
$
L_{det} = L_{2D} + L_{1D} + L_{\mathrm{bbox}}
$.
\subsubsection{Instantaneous Pose Estimation}

For each detection $D$, we construct a volume of fixed size centered on the detection's center $c_i$, and unproject the backbone features from each camera view into the volume, as in the detection step. As in \cite{fastervoxelpose}, we mask out all features falling outside the detection's associated bounding box $B_i$, resulting in a feature volume $V^t_i$ for person $i$.

As shown in Figure \ref{fig:temporal_rep_figure}, we project the feature volume $V^i_t$ to 2D along each of the $xy, yz,$ and $xz$ planes, resulting in three 2D feature maps, denoted by $\mathbf{P}^t_{i, xy}$, $\mathbf{P}^t_{i, xz}$, and $\mathbf{P}^t_{i, yz}$. 
The intuition behind this step is that we can predict the 2D position of each joint in each plane, and fuse the predicted 2D positions back together to form a 3D skeleton. 
Each feature map is passed through a 2D CNN to decode a heatmap of joint likelihood for every person joint, in each of the three planes, and the 2D joint predictions from each plane are fused into 3D with a learned weighting network. 
We define both the loss for a predicted pose as the mean squared loss between the computed and ground truth 2D heatmaps, as well as the $L_1$ loss of the predicted joint locations and ground truth:
\begin{equation}
\begin{aligned}
L^k_{joint, t} = \mathrm{MSE}(\mathbf{P}^t_{xy}, \mathbf{P}^t_{xy, GT}) + \mathrm{MSE}(\mathbf{P}^t_{xz}, \mathbf{P}^t_{xz, GT}) + \\ 
\mathrm{MSE}(\mathbf{P}^t_{yz}, \mathbf{P}^t_{yz, GT}) + \sum_{j=1}^J | j_{i, pred} - j_{i, GT}|
\end{aligned}
\end{equation}
 with $\mathrm{MSE}$ representing the mean squared error. 

\subsection{Person Tracking}
We now describe how TEMPO uses temporal information. Unlike previous works, TEMPO takes as input a set of $N$ images at time $t$ as well as the person detections $D_{t-1}$ and corresponding 2D pose embeddings $P^{t-1}$ from the previous timestep.

Each proposed detection from the previous step consists of a body center $c_i$ and a bounding box $B^t_i = (h^t_i, w^t_i)$. Given $K$ proposals, we compute a $K \times K$ cost matrix $\mathbf{A}$ based on the IoU between $B^t_i$ and $B^t_k$ for all detections at time $t$, resulting in 
\begin{equation}
    \mathbf{A}[i][j] = \lVert{c_i - c_j}\rVert
\end{equation} with $c_i, c_j$ being the associated predicted locations of person $i$ and person $j$'s root joint. 

While VoxelTrack \cite{zhang2022voxeltrack} computes cost across every joint in each pose, TEMPO uses the top-down view of the associated bounding box for each person. At inference time, we use the SORT \cite{bewley2016simple} tracker, which is fast and uses a simple Kalman filter with no learned Re-ID mechanism. While \cite{tessetrack} uses a learned tracker based on SuperGlue \cite{sarlin20superglue}, we find that SORT is faster and does not result in degraded performance.

\subsection{Temporal Pose Estimation and Forecasting}

After running the tracker, the input to the pose estimation stage is a set of detection proposals $D^t$, the previous timestep's detection proposals $D^{t-1}$, and the assignment matrix between the detection sets $\mathbf{A}^{t}$. We also assume access to pose features $\mathbf{P}^{t-1}_i$ for each person in the previous timestep. Both $D^t$ and $D^{t-1}$ have $K$ proposals each.

However, $\mathbf{P}^{t-1}_i$ and $\mathbf{P}^t_i$ are centered at $c^{t-1}_i$ and $c^{t}_i$ respectively, and thus the pose features from each volume are not in the same coordinate system due to the motion of person $i$. To fix this, we follow the standard procedure used in temporal birds-eye view prediction \cite{bevformer, fiery} and warp the projected features from $\mathbf{F}^{t-1}_i$ into the coordinate system of $\mathbf{P}^t_i$ with a translational warp defined by $c^{t}_i - c^{t-1}_i$.

After warping the previous pose features, we run a recurrent network with Spatial Gated Recurrent Units \cite{ballas2015delving} (SpatialGRUs) to produce multiple embeddings: $\mathbf{F}^t_i$, representing the current pose, and $\mathbf{F}^{t+1}_i, \mathbf{F}^{t+2}_i, \hdots$, representing the pose in future frames.
 While \cite{fiery} and \cite{bevformer} do not propagate gradients through time and only predict object locations and instance segments at time $T$, we \textit{do} backpropagate through time by predicting the pose for each person at \textit{every} timestep. At training time, we recurrently compute the temporal representation at each timestep $t_0, t_0 + 1, \hdots t_0 + T$, decode a pose for every timestep, and compute losses over all the predicted poses simultaneously. Thus, the final training objective is 
\begin{equation}
L_{\mathrm{pose}} = \sum_{t=1}^T \sum_{i=1}^K L^i_{\mathrm{joint}, t} + L^i_{\mathrm{joint}, t+1}
\end{equation}
where $L^i_{\mathrm{joint}, t}$ is the L1 distance between the predictde and ground truth pose at time $t$
Providing supervision to the network at every timestep allows the network to learn a representation that encodes the motion between consecutive frames while enabling temporal smoothness between predictions. As we show in Section \ref{sec:ablations}, this technique is crucial to our model's performance.

While training, we run the network $T$ times, for each input timestep. However, at inference time, we save the previous embeddings and only receive input images at a single timestep $t$, significantly reducing the computational burden and allowing our model to use temporal information without sacrificing efficiency.

\begin{table*}
\begin{center}
\begin{tabular}{llcccccccc}
\toprule
Method & Backbone & Resolution & $\mathrm{AP}_{25}\uparrow$ & $\mathrm{AP}_{50}\uparrow $& $\mathrm{AP}_{100}\uparrow$ & $\mathrm{AP}_{150}\uparrow$ & MPJPE (mm) $\downarrow$ & FPS (s)$\uparrow$ \\
\midrule
VoxelPose\cite{voxelpose} & ResNet-50 & $960 \times 512$ & 83.59	&98.33	&99.76	&99.91	&17.68 & 3.2 \\
Faster VoxelPose \cite{fastervoxelpose} & ResNet-50 & $960 \times 512$ & 85.22 & 98.08 & 99.32 & 99.48 & 18.26 & \textbf{31.1} \\
PlaneSweepPose \cite{planesweeppose} & ResNet-50 & $960\times512$ & 92.12	& 98.96	&99.81	&99.84	&16.75 & 4.3 \\
MvP \cite{planesweeppose} & ResNet-50 & $960 \times 512$ & \textbf{92.28}	& 96.6	& 97.45 & 97.69 & 15.76 & 3.6 \\
Ours & ResNet-50 & $960\times512$ & 89.01 & \textbf{99.08} & \textbf{99.76} & \textbf{99.93} & \textbf{14.68} & 29.3 \\
\midrule
VoxelPose\cite{voxelpose} & HRNet & $384 \times 384$ & 82.44 &  \textbf{98.55} & 99.74& 99.92& 17.63 &  2.3 \\
Faster VoxelPose \cite{fastervoxelpose}& HRNet & $384 \times 384$ & 81.69& 98.38& 99.67& 99.83 & 18.77&  22.4 \\
MvP \cite{wang2021mvp} & HRNet & $384\times384$ & \textbf{90.41} & 96.32 & 97.39 & 97.89 & 16.34 & 2.8\\
TesseTrack$^\dagger$ \cite{tessetrack} & HRNet & $384\times384$ & 86.24 & 98.29 &  99.72& 99.50 & 16.92 & 0.6 \\
Ours & HRNet & $384\times384$ & 89.32 &  98.48 &  \textbf{99.73} &  \textbf{99.94} & \textbf{15.99} & 20.3 \\
\bottomrule
\end{tabular}
\end{center}
\caption{Pose estimation results on the CMU Panoptic dataset. Our method achieves the best MPJPE and AP while running at speed comparable to Faster VoxelPose. We evaluate our methods at $384\times384$ resolution on the Panoptic dataset, as well as the higher resolution used in other methods. We mark TesseTrack \cite{tessetrack} with a $\dagger$ as their reported results are on a different data split, and the results in this table are from our best reproduction of the method, which is not public.}
\label{Tab:pose_estimation_panoptic}
\end{table*}\
\begin{table*}
\begin{center}
\begin{tabular}{lcccccccc}
\toprule
& \multicolumn{4}{c@{}}{Shelf} & \multicolumn{4}{c@{}}{Campus}\\
\cmidrule(l){2-5} \cmidrule(l){6-9}
Method & Actor-1 & Actor-2 & Actor-3 & Average & Actor-1 & Actor2 & Actor3 & Average \\
\cmidrule(l){1-1} \cmidrule(l){2-5} \cmidrule(l){6-9}
Belagiannis et al. \cite{belagiannis_campus} & 66.1 & 65.0 & 83.2  & 71.4 & 82.0 & 72.4 & 73.7 & 75.8  \\
Ershadi et al. \cite{ershadi2018multiple} & 93.3 & 75.9 & 94.8 & 88.0 &  94.2 & 92.9 & 84.6 & 90.6 \\
Dong et al.\cite{dong2019fast} & 98.8 & 94.1 & 97.8 & 96.9 & 97.6 & 93.3 & 98.0 & 96.3 \\
VoxelPose \cite{voxelpose} &  99.3 & 94.1 & 97.6 & 97.0 &  97.6 & 93.8 & 98.8 & 96.7 \\
Faster VoxelPose\cite{fastervoxelpose} & 99.4 & 96.0 & 97.5 & 97.6 &  96.5 & 94.1 & 97.9 & 96.2 \\
PlaneSweepPose\cite{planesweeppose} &  99.3 &  \textbf{96.5} & 98.0 & 97.9 & 98.4 & 93.7 & \textbf{99.0} & 97.0 \\ 
MvP \cite{wang2021mvp} & 99.3 & 95.1 & 97.8 & 97.4 &  98.2 & 94.1 & 97.4 & 96.6 \\ 
\textbf{TEMPO} (Ours) & 99.0 &  96.3 &  \textbf{98.2} &  \textbf{98.0} & 97.7 & \textbf{95.5} & 97.9 & \textbf{97.3} \\
\bottomrule
\end{tabular}
\end{center}
\caption{PCP3D accuracy on the Campus and Shelf datasets. We follow the protocol of previous methods and train our backbone on synthetic heatmaps of ground-truth poses. Our method achieves results comparable to the state-of-the-art.}
\label{tab:campus_shelf_pose}
\end{table*}
\section{Experiments}
\subsection{Datasets and Metrics}
\textbf{Panoptic Studio} The CMU Panoptic Studio dataset \cite{joo_iccv_2015} is a large multi-view pose estimation dataset with several synchronized camera sequences of multiple interacting subjects. Following prior work \cite{voxelpose, fastervoxelpose,wang2021mvp, planesweeppose}, we use five HD cameras, specifically cameras 3, 6, 12, 13, and 23. We also use the same training and test split as these works, omitting the sequence \texttt{160906\_band3} due to data corruption. 

\textbf{Human 3.6M} The Human3.6M dataset \cite{h36m_pami, IonescuSminchisescu11} consists of videos of a single subject in an indoor studio with four static cameras. Each video has a professional actor performing a specific action. We follow the training-test split of prior works, using subjects 9 and 11 for validation and the others for training, while omitting corrupted sequences. 

\textbf{Campus and Shelf} The Campus and Shelf datasets\cite{belagiannis_campus} contain approximately 4000 frames of a single scene. While these datasets are commonly used for benchmarking in previous work, they are missing many annotations. We follow previous work \cite{voxelpose, fastervoxelpose, wang2021mvp} and adopt a synthetic heatmap-based scheme for training.

\textbf{EgoHumans} We include the newly collected EgoHumans multi-view dataset \cite{khirodkar2023egohumans}. This benchmark consists of approximately 1 hour of video of up to 5 subjects performing highly dynamic activites, such as playing tag, fencing, or group assembly. It contains videos from up to eight fisheye cameras and includes both egocentric and exocentric camera and pose data.

\textbf{Metrics} For the Panoptic Studio, Human3.6M, and EgoHumans datasets, we report the mean per-joint position error (MPJPE).
We additionally report the Average Precision ($\mathrm{AP}_K$) on the Panoptic and EgoHumans datasets and report MPJPE for Human3.6M in line with previous work \cite{pavlakos2017harvesting, martinez20173dbaseline, iskakov2019learnable, tessetrack}. On the Shelf and Campus dataset we report the Percentage of Correct Parts (PCP3D). For pose forecasting, we measure the MPJPE between the predicted pose and the ground truth pose 0.33s into the future, matching previous work \cite{snipper}.

\begin{table*}[htb]
\begin{subtable}[t]{.35\textwidth}
\setlength{\tabcolsep}{1.5pt}
\begin{center}
\begin{tabular}{lccc}
\toprule
Method &  MOTA & IDF1 &  MPJPE \\
\midrule
VoxelTrack \cite{zhang2022voxeltrack} & 98.45 & 98.67 & - \\ 
Snipper \cite{snipper} & 93.40 & 85.50  & 40.2 \\
\textbf{TEMPO} (Ours) & 98.42 & 93.62 & \textbf{38.5} \\
\bottomrule
\end{tabular}
\end{center}
\caption{Evaluation of tracking and forecasting on the CMU Panoptic dataset. Our method outperforms \cite{snipper} by using multi-view information and is competitive with VoxelTrack. \cite{zhang2022voxeltrack}}
\label{tab:forecast_eval}
\end{subtable}
\hspace{1cm}
\begin{subtable}[t]{.6\textwidth}

\centering
\setlength{\tabcolsep}{1pt}
\begin{tabular}{ccc|cccc}

\toprule
& Training Dataset & & & MPJPE (mm) & \\
\midrule
Panoptic & H.6M & EgoHumans  & Panoptic & H.6M & EgoHumans\\
\midrule
\checkmark & & & 14.68 & 62.96 & 119.8\\
\checkmark & \checkmark & & 36.56 &  25.3 & 108.3 \\
\checkmark & \checkmark & \checkmark & 42.8 & 32.3 & 48.9\\
\bottomrule
\end{tabular}

\caption{Evaluating TEMPO's ability to transfer. Unlike previous methods, TEMPO is able to train on multiple datasets and can perform reasonably given enough training data. While our model is able to effectively transfer to Human3.6M, it has difficulty with EgoHumans due likely due to its use of fisheye cameras.}
\label{tab:transfer_pose}
\end{subtable}
\end{table*}

\begin{table}
\begin{center}
\begin{tabular}{lc|c}
\toprule
Method & \rotatebox[origin=c]{90}{Human 3.6M}  & \rotatebox[origin=c]{90}{EgoHumans}\\
\midrule
Martinez et al. \cite{martinez20173dbaseline} & 57.0 & - \\
Pavlakos et al. \cite{pavlakos2017harvesting} & 56.9  & -\\
Kadkhodamohammadi et al. \cite{kadkhodamohammadi2021generalizable} & 49.1  & -\\
Ma et al \cite{ma2022ppt} & 24.4 & -\\
VoxelPose \cite{voxelpose} & 19.0  & 46.24 \\ 
Faster VoxelPose \cite{fastervoxelpose} & 19.8 & 48.50 \\
MvP \cite{wang2021mvp} & 18.6 & 41.72 \\
Iskakov et al. \cite{iskakov2019learnable} & \textbf{17.7} & -  \\
\textbf{TEMPO} (Ours) & 18.5 & \textbf{36.74} \\
\bottomrule
\end{tabular}
\end{center}
\caption{Pose Estimation results on the Human3.6M and EgoHumans datasets in MPJPE (mm). Our method is competitive woth the state-of-the art on Human3.6M and surpasses current methods by a significant margin on EgoHumans.}
\label{tab:egohumans_h36m_pose}
\end{table}

\subsection{Implementation Details}
\vspace{-0.5em}
Following \cite{voxelpose, fastervoxelpose, snipper, wang2021mvp} we use a ResNet-50 backbone pre-trained on the Panoptic dataset and follow \cite{iskakov2019learnable} by also using a ResNet-50 backbone pre-trained on Human3.6M for the dataset-specific pose estimation results.
We use HRNet \cite{hrnet} pre-trained on COCO as the model backbone for the generalization experiments, as done in \cite{tessetrack} rather than pose estimation backbones that are trained on multi-view datasets.
All methods are trained on 8 NVIDIA A100 GPUs with batch size of 2 per GPU. We use Adam with a learning rate of 3e-4, with weight decay of 1e-4 and a linear decay schedule for 10 epochs. We measure FPS using a single A100 GPU, and our code is based off the MMPose \cite{mmpose2020} library. Additional architectural details are in the supplement. 

\subsection{Pose Estimation Results}
\vspace{-0.5em}

We first compare the results of our methods other state-of-the art methods. On the Panoptic Studio dataset, we report results following \cite{fastervoxelpose, voxelpose, wang2021mvp} and use $960\times512$ images and initialize from a ResNet-50 \cite{resnet, xiao2018simple} checkpoint pretrained on the Panoptic dataset. We also evaluate our method using an HRNet \cite{hrnet} backbone pretrained on COCO with $384\times384$ images for a fair comparison with TesseTrack.
In Table \ref{Tab:pose_estimation_panoptic} we provide a complete comparison across the state-of-the-art methods, with baselines trained on the same image resolutions and backbones for completeness. TEMPO achieves significantly lower MPJPE across both resolutions and backbones, while running at 29.3 FPS, competitive with Faster VoxelPose. We attribute this performance increase to the the smoother and more temporally consistent skeletons our model produces due to its incorporation of temporal context and temporal supervision. We also show  in \ref{tab:campus_shelf_pose} that our model achieves performance competitive with the state of the art on the Campus and Shelf datasets.

\begin{figure*}
    \centering
    \includegraphics[width=\linewidth]{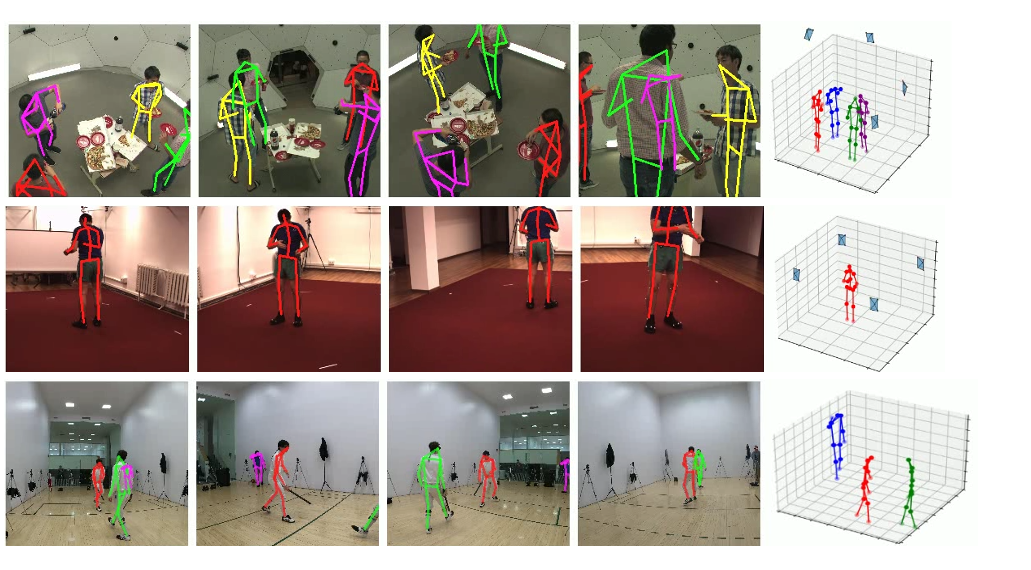}
    \caption{Samples of our model's pose estimation performance on the Panoptic Studio, Human3.6M, and EgoHumans datasets. TEMPO predicts accurate poses and tracks them over time.}
    \label{fig:sample_results_tile}
\end{figure*}
We compare pose estimation performance on the Human3.6M  and EgoHumans datasets in Table \ref{tab:egohumans_h36m_pose} . Our results perform comparably to the state-of-the art. Notably \cite{iskakov2019learnable} uses ground-truth locations and cropped bounding boxes from input views, while our method is able to match performance despite simultaneously detecting people in the scene. Furthermore, our method significantly outperforms others in the more challenging EgoHumans benchmark, suggesting that temporal context is crucial for handling rapid motion.

\subsection{Pose Tracking and Forecasting}
 We compared the performance of TEMPO's tracking to VoxelTrack \cite{zhang2022voxeltrack} in Table \ref{tab:forecast_eval}. Our tracker is competitive but performs slightly worse, which is expected due to its lack of learned Re-ID features. 
 To our knowledge, pose forecasting has not been attempted for the multi-view case, so we compare against the closest forecasting method, Snipper \cite{snipper}, which estimates future pose from monocular video. 
Our method takes as input 4 previous timesteps and predicts the pose over the next 3 timesteps, which is 0.33 seconds into the future, matching prior work \cite{snipper}. Our model achieves state-of-the-art performance, shown in Table \ref{tab:forecast_eval}.

We conducted an additional experiment to measure the performance of our model in transfer on different datasets. The standard practice in monocular 3D pose estimation is to train on and evaluate performance on a combination of multiple datasets. However, the predominant paradigm for evaluating multi-view pose estimation methods has been to train a single  model only on a single dataset and evaluate it on the same dataset.This severely limits the potential of these models to generalize across different settings and thus be deployed in real-world scenarios. Similar to \cite{iskakov2019learnable}, we evaluate the performance of our model trained on multiple combinations of annotated datasets, and report the results for each combination in Table \ref{tab:transfer_pose}. We run our model with no fine tuning and the same voxel size  of 10cm$^3$ across each dataset. Our method is able to transfer and successfully tracks and predicts pose, but performs noticeably worse, likely due to being trained on a single camera configuration. In particular, the CMU Panoptic training dataset uses 5 cameras, whereas Human3.6M uses 4 and EgoHumans uses 8 fisheye cameras. The model has the most trouble generalizing to EgoHumans, likely due to the significantly larger indoor space and different camera models. We find that TEMPO performs better in transfer after including training data from each dataset, especially on the EgoHumans dataset, suggesting that future works should include diverse multi-view data from different camera configurations and camera models in order to better generalize.

\subsection{Ablations}
\begin{table}
    \label{tab:ablations}
    \centering
    \setlength{\tabcolsep}{2pt}
    \begin{tabular}{l|ccccc}
    \toprule
    Method & $T$ & Forecasting & Warping & Per-$t$ loss &  MPJPE $\downarrow$(mm) \\
    \midrule
    (a) & 3 & & & & 17.83  \\
    (b) & 3 & & & \checkmark & 15.03 \\
    (c) & 3 & & \checkmark & \checkmark & 14.94 \\
    (d) & 3 & \checkmark & \checkmark & \checkmark  & 14.68 \\
    (e) & 4 & \checkmark & \checkmark & \checkmark  & 14.90 \\
    (f) & 5 & \checkmark & \checkmark & \checkmark  & 14.82 \\
    \bottomrule
    \end{tabular}
    \caption{Ablation study on various components to our model. The most important component to performance was the per-timestep supervision. Warping the previous feature also improved performance. We observed the forecasting and slightly increasing the length of the input history had no noticeable effect on performance.}
    \label{tab:ablation}
\end{table}
\begin{table}[t]
\centering
\begin{tabular}{l|ccccc}
\toprule
Cameras &  1 & 2 & 3 & 4 & 5 \\
MPJPE & 51.32 & 32.13 & 19.22 & 17.34 & 14.68 \\
\bottomrule
\end{tabular}
\caption{Ablation on number of cameras. We observe that performance decreases with decreasing number of cameras.}
\label{tab:cam_ablation}
\end{table}
\vspace{-0.5em}
\label{sec:ablations}
We train ablated models to study the impact of individual components in our method. All our experiments are conducted on the CMU Panoptic dataset, with results shown in table \ref{tab:ablation}. We find that using temporal information slightly helps, but with per-timestep supervision, the model is able to greatly improve. Warping the pose between timesteps further improves the model. We hypothesize that the improvement from warping is small due to the relative lack of motion in the Panoptic dataset - the distance between body centers in consecutive timesteps is usually small. We also measure the effect of the history length on performance and found no significant difference. While intuitively, larger history lengths should provide more context, the GPU memory constraints of our method prevent investigating $T > 5$. We also ablated on the number of cameras on the Panoptic dataset. We found that the MPJPE increases with the number of cameras, matching the findings of \cite{voxelpose, iskakov2019learnable} and \cite{tessetrack}.

\section{Conclusions}
Understanding human behavior from video is a fundamentally temporal problem, requiring accurate and efficient pose estimation algorithms that can reason over time. We presented the first method that satisfies these requirements, achieving state-of-the-art results over existing multi-view pose estimation benchmarks via temporal consistency as a learning objective. Our model is also highly efficient, relying on recurrence to maintain a temporal state while enabling pose tracking and forecasting. TEMPO represents a step closer towards general-purpose human behavior understanding from video.

\section*{Acknowledgements}
This research was supported partially by Fujitsu. 

{\small
\bibliographystyle{ieee_fullname}
\bibliography{egbib}
}

\clearpage

\appendix

\section{Implementation Details}

Our code is based on the MMPose\cite{mmpose2020} public repository , and we used their built-in implementations for the image backbone as well as the inference time analysis tools.

\textbf{Backbone} We use ResNet-50 \cite{xiao2018simple, resnet} as our backbone. On the Panoptic studio dataset, we use the checkpoint trained for 20 epochs on the Panoptic Studio dataset with $960 \times 512$ resolution images, introduced by the VoxelPose\cite{voxelpose} codebase for accurate comparison with existing methods. Since we use synthetic heatmaps for Shelf and Campus, we use no backbone. On Human3.6M, we use the pre-trained ResNet backbone from the Learnable Triangulation \cite{iskakov2019learnable} codebase.
On all other datasets, we used HRNet\cite{hrnet} with $384 \times 384$ resolution, with no pre-training, following TesseTrack\cite{tessetrack}.
Following MvP \cite{wang2021mvp}, we use the pre-final layer of the backbone model's output head rather than the final per-joint heatmaps. This pre-final layer has 256 channels for ResNet and 32 for HRNet.
\begin{figure*}
    \centering
    \includegraphics[width=0.9\linewidth]{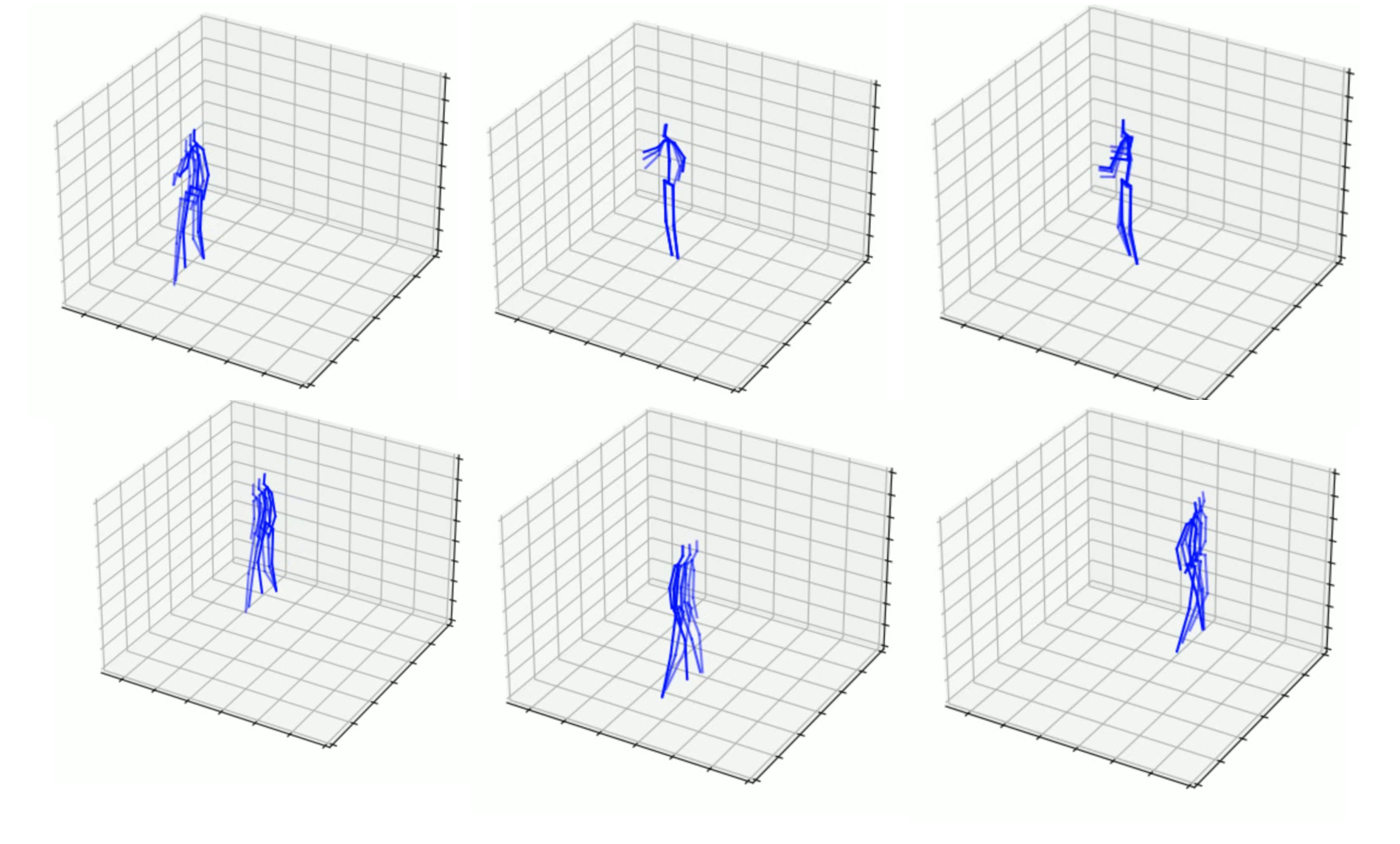}
    \caption{Sample forecasting outputs on the Human3.6M dataset. Our model produces feasible forecasts up to 0.33 seconds into the future, surpassing the accuracy of comparable works \cite{zou2023snipper}.}
    \label{fig:forecasting}
\end{figure*}
\textbf{Detector} The person detector follows the design of \cite{fastervoxelpose}. We used a fixed voxel size of 10 $\mathrm{cm}^3$. For dataset-specific training, we follow previous papers and used a volume size of  of $80 \times 80 \times 20$,  on the Panoptic Studio dataset. We use the basic structure of V2V-Net \cite{moon2018v2v}, for the networks in this stage, but in 2D and 1D. The building block of this network consists of a convolutional block and a residual (skip-connection) block, with a ReLU connection and BatchNorm. We first feed the input to the network through a layer with a $7 \times 7$ kernel and then passed through three successive blocks, each with $3 \times 3 $ kernels, with maxpooling between each block with kernel size 2. We then apply three transposed convolutional layers to obtain a feature map with the same spatial size as the input, to which we apply a $1 \times 1$ convolution to get the desired channel output size.

\begin{table}
    \centering
    \begin{tabular}{lccc}
    \toprule
       Component  &  Time (ms) & GFLOPs & Params \\
       \midrule
       Backbone & 11.72 & 29.3 & 23.51M \\
       \midrule
       Detector (FV) & 17.3 & 1.204 & 1.51M\\
       Detector(Ours) & 17.3 & 1.204 & 1.51M \\
       \midrule 
       Pose (FV) & 14.9& 6.621 & 1.13M\\
       Pose (Ours) & 16.5& 7.331 & 1.926M\\
       \midrule
       Total (FV) & 43.92& 37.125& 32.40M \\
       \textbf{Total} & 45.52 & 38.831 & 33.19M\\ 
       \bottomrule
    \end{tabular}
    \caption{Runtime analysis of TEMPO compared with Faster VoxelPose (FV) \cite{fastervoxelpose}. Our model is competitive with Faster VoxelPose, which is the state-of-the-art in efficiency. Our model achieves significantly better pose estimation performance despite adding relatively few parameters and without adding significant overhead.}
    \label{tab:runtime_breakdown}
\end{table}

\subsection{Cross-dataset Generalization}
\begin{table}[t]
\centering
\begin{tabular}{lccc}
\toprule
Method & Panoptic & Human3.6M\\ 
\midrule
MVPose & 55.6 & 83.4\\
VoxelPose & 17.68 & 273.2\\
Faster VoxelPose & 18.26 & 283.1\\
TEMPO (Ours) & 14.18 & 63.4 & \\
\bottomrule
\end{tabular}
\caption{TEMPO significantly surpasses optimization-based methods on datasets it was not trained on, despite their dataset-agnostic design}
\label{tab:generalization}
\end{table}

\textbf{Pose Estimation and Forecasting}
The recurrent network we used was based on the SpatialGRU implementation used in FIERY \cite{fiery} with a 2D LayerNorm based on the official ConvNexT implementation \cite{liu2022convnet}.

At each timestep, the 2D projected features were fed into an encoder with the same structure as the encoder portion of the 2D CNN used in the detection stage. 
We then feed the encoded features through the RNN, and run a 2D CNN with the same structure as the detection network's decoder. on the hidden state output. The output of the decoder network was fed into a learned weight network with the exact same structure as in Faster VoxelPose \cite{fastervoxelpose}.

We used 4 timesteps of input at training time, following the augmentation scheme of BEVFormer, and the forecasting output is 2 timesteps into the future, each 3 frames apart. At inference time, we only feed a single timestep of input into the network, and TEMPO saves the previous embedding features, matching them to detections at each timestep with the tracker.

\textbf{Training Details}
For the ResNet backbone, we trained the entire network jointly to convergence for 10 epochs with a batch size of 1. We used the Adam optimizer with weight decay 1e-4, learning rate 1e-4, and applied a linear decay schedule with $\gamma = 0.7$, updating every 2 epochs.
We used a batch size of 2 and trained the network for 20 epochs, and used a learning rate of 5e-4, with all other parameters the same.
For the Panoptic, Human3.6M, and DynAct datasets, we used images as input, while for the Shelf and Campus dataset we followed the scheme of \cite{voxelpose, fastervoxelpose, tessetrack, wang2021mvp} and used synthetic joint heatmaps, produced by projecting ground-truth poses from the Panoptic dataset onto the cameras in the Campus and Shelf dataset. 

\section{Additional Ablation Details}
\subsection{Cross-dataset Generalization}
Although TEMPO is not explicitly designed to provide strong generalization across multi-view datasets, we found that simply computing the space and volume dimensions from the camera configuration, it was able to transfer surprisingly well. In Table \ref{tab:generalization}, we show that TEMPO exceeds both VoxelPose and Faster VoxelPose in this regard. Furthermore, TEMPO significantly exceeds the performance of MVPose \cite{dong2019fast}, a method that is based on graph optimization and is dataset-agnostic by design, underscoring the strength of volumetric pose estimation methods.

\subsection{Inference Time}
We conducted a more detailed inference time analysis, comparing our work with Faster VoxelPose \cite{fastervoxelpose}, the current fastest method. Our results are shown in Table \ref{tab:runtime_breakdown}. In the main text, we follow the convention of \cite{fastervoxelpose, wang2021mvp, planesweeppose} and omit the runtime of the image backbone. We include it here for a full picture of our method's speed. Since the image backbone time is dependent on the number of views, we used 5 views for testing, in line with the Panoptic Studio dataset. We benchmarked all our models on a Nvidia A-100 with a AMD EPYC 7352 24-Core Processor @ 2.3GHz CPU.

For each module in our model, we provide the inference time, GFLOPs, and number of parameters. Since both ours and Faster VoxelPose are top-down methods, the GFLOPs and runtime vary with the number of detections. In this analysis, we used 3 detections for both methods.

\end{document}